\documentclass[10pt,twocolumn,letterpaper]{article}

\usepackage[pagenumbers]{cvpr} 

\newcommand\blfootnote[1]{%
  \begingroup
  \renewcommand\thefootnote{}\footnote{#1}%
  \addtocounter{footnote}{-1}%
  \endgroup
}

\definecolor{cvprblue}{rgb}{0.21,0.49,0.74}
\usepackage[pagebackref,breaklinks,colorlinks,allcolors=cvprblue]{hyperref}

\def\paperID{*****} 
\def\confName{CVPR}
\def\confYear{2025}

\usepackage{amsmath,amsfonts,bm}

\newcommand{\rvx}[0]{\mathbf{x}}

\newcommand{\mA}[0]{\bm{A}}

\newcommand{\mD}[0]{\bm{D}}

\newcommand{\mK}[0]{\bm{K}}

\newcommand{\mT}[0]{\bm{T}}

\usepackage{amsmath}

\definecolor{darkgreen}{rgb}{0.0, 0.5, 0.13}

\title{Novel View Synthesis with Pixel-Space Diffusion Models}

\author{Noam Elata$^*$\\
Technion\\
{\tt\small noamelata@campus.technion.ac.il}
\and
Bahjat Kawar\\
Apple\\
{\tt\small b\_kawar@apple.com}
\and
Yaron Ostrovsky-Berman\\
Apple\\
{\tt\small yostrovskyberman@apple.com}
\and
Miriam Farber\\
Apple\\
{\tt\small m\_farber@apple.com}
\and
Ron Sokolovsky\\
Apple\\
{\tt\small ronus@apple.com}
}

\begin{document}
\maketitle
\blfootnote{$^*$ Work done as part of an internship at Apple.}

\begin{abstract}
Synthesizing a novel view from a single input image is a challenging task.
Traditionally, this task was approached by estimating scene depth, warping, and inpainting, with machine learning models enabling parts of the pipeline.
More recently, generative models are being increasingly employed in novel view synthesis (NVS), often encompassing the entire end-to-end system.
In this work, we adapt a modern diffusion model architecture for end-to-end NVS in the pixel space, substantially outperforming previous state-of-the-art (SOTA) techniques.
We explore different ways to encode geometric information into the network.
Our experiments show that while these methods may enhance performance, their impact is minor compared to utilizing improved generative models.
Moreover, we introduce a novel NVS training scheme that utilizes single-view datasets, capitalizing on their relative abundance compared to their multi-view counterparts.
This leads to improved generalization capabilities to scenes with out-of-domain content.
\end{abstract}

\section{Introduction}

\begin{figure*}
    \centering
    \includegraphics[width=\linewidth]{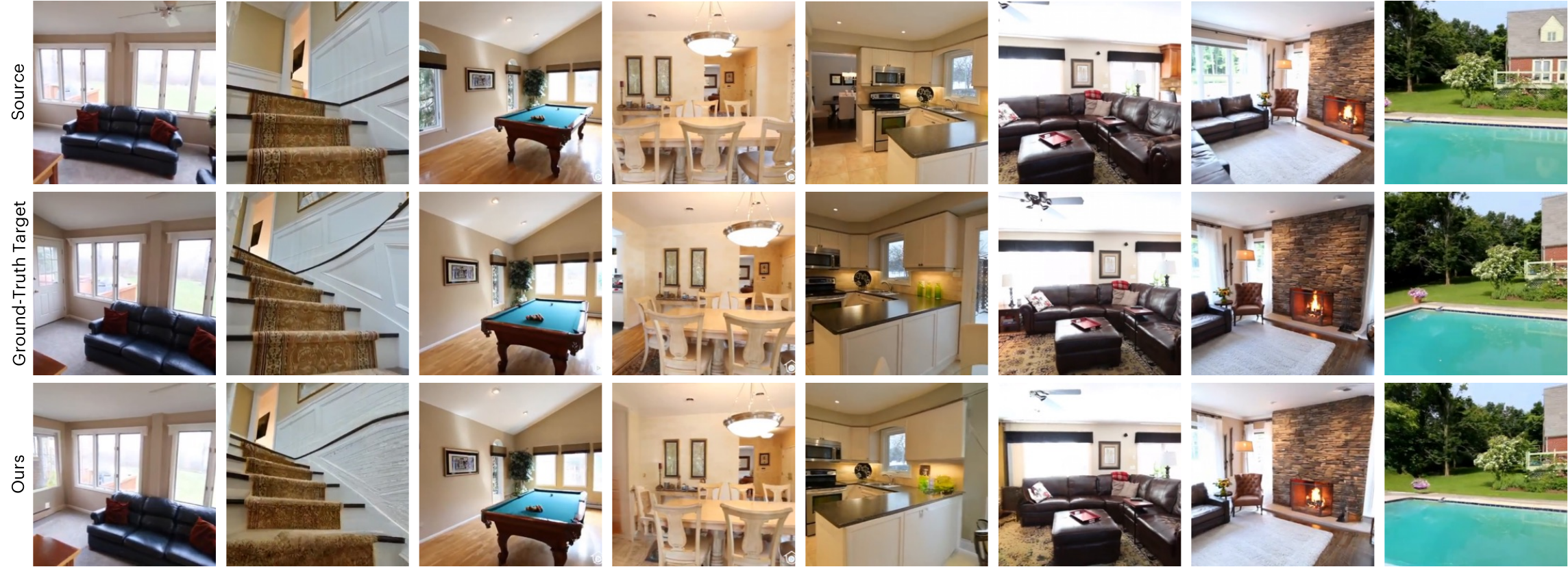}
    \caption{Novel view synthesis results from our diffusion model. Source views are taken from RealEstate10K~\cite{realestate10k}, and fed into our base and SR models to produce a $256\times256$-pixel prediction. Our end-to-end system implicitly learns to preserve the features in the source view, transform their position along with the camera movement, and generate realistic details in unseen areas.}
    \label{fig:results}
\end{figure*}
In novel view synthesis (NVS), we aim to recreate a snapshot of a given scene from an unseen perspective.
A successful algorithm must consider both the 3D geometry of the given scene and the underlying distribution of real-world images.
NVS algorithms have been researched for many years, covering several tasks which differ in the expected number of input and output views and the relative distance between inputs and outputs, among other things.
In this work, we focus on the simplest form of NVS: single-image to single-image, hoping that it will serve as a foundation for more general NVS settings.
In previous works, this task has been decomposed into a pipeline of several computer vision components, namely depth estimation, warping, and inpainting~\cite{wiles2020synsin, shih20203d, liu2021infinite, chung2023luciddreamer}.
However, modern advances in generative modeling offer a simpler and more robust end-to-end approach~\cite{rombach2021geometry, poseguideddiffusion, yu2023long, seo2024genwarp}.

Generative diffusion models~\cite{ho2020denoising, song2019generative, sohl2015deep} have emerged as a top class of image generators, excelling at many conditional generation tasks~\cite{kawar2022denoising, saharia2022photorealistic}, including NVS~\cite{yu2023long, poseguideddiffusion, seo2024genwarp}.
In diffusion modeling, we train a neural network for removing Gaussian noise, and iteratively apply it in several steps to generate an image.
These models are relatively simple to train, owing their stability to a simple denoising regression loss.
Modern diffusion architectures notably rely on transformer layers~\cite{vaswani2017attention} with self- and cross- attention blocks.
These attention mechanisms are highly effective for the conditional generation task.
For instance, in the widely used text-to-image generators~\cite{saharia2022photorealistic, rombach2022high}, cross-attention is used to condition the denoising network's features on the textual tokens.
In our work, we effectively harness cross-attention in diffusion models for the NVS task.

Some existing diffusion-based NVS works~\cite{gao2024cat3d, seo2024genwarp} operate in the latent space of an auto-encoder, following~\cite{rombach2022high}.
This can result in an unnecessary loss of high-frequency details due to the auto-encoder's reconstruction error, leading to texture mismatches between source and target views.
Therefore, we opt to apply our method directly in the pixel space using a cascaded diffusion model design~\cite{ho2022cascaded}, avoiding this issue altogether.
We demonstrate the superior texture transfer capabilities of our model compared to latent space NVS models.

Furthermore, many recent generative model-based NVS works involve conditional diffusion training with some form of intricate 3D geometry encoding~\cite{yu2023long, poseguideddiffusion, seo2024genwarp}.
However, in many cases, the benefit of such complex encoding methods is unclear, with some researchers claiming they may not be needed at all~\cite{rombach2021geometry}.
Inspired by previous work, we explore geometry encoding methods and perform an ablation study.
In our experiments, we show that that the impact of these methods is minimal, especially when harnessing a powerful diffusion model architecture.
Building on our conclusions, we train an NVS diffusion model, reaching state-of-the-art (SOTA) NVS capabilities for the commonly accepted RealEstate10K dataset~\cite{realestate10k}. Our model, which we call VIVID (\textbf{V}iew \textbf{I}nference \textbf{V}ia \textbf{I}mage \textbf{D}iffusion), excels in both image quality and fidelity to the ground-truth novel view, measured by FID~\cite{fid} and PSNR, respectively.

Finally, we experiment with the generalization capabilities of our method, and attempt to overcome the limited availability of multi-view data.
Using the insight that camera rotations can be accurately simulated with a simple 2D homography transform, we propose a novel data augmentation scheme that enables the use of single-view datasets in NVS training.
Our proposed scheme unlocks the possibility of training NVS models on far richer image content distributions, without introducing data scale mismatches that commonly occur in multi-dataset NVS training~\cite{sargent2024zeronvs}.
We show the effectiveness of this scheme in generalizing to unseen scenes with out-of-distribution content relative to the multi-view training dataset.

To summarize, we consider the task of novel view synthesis (NVS), characterize its different flavors, and focus on single-image to single-image NVS.
We make use of a powerful diffusion model backbone~\cite{karras2024analyzing}, adapt it for NVS using the cross-attention layers, and apply it in the pixel space rather than a latent one.
We ablate on different geometry encoding methods, and conclude that they offer minimal improvement over a simple scalar embedding of the camera poses.
Our resulting model, called VIVID, achieves state-of-the-art NVS performance on the widely accepted RealEstate10K~\cite{realestate10k} benchmark.
However, our method has some limitations that we discuss in \autoref{sec:conclusion}, which we hope to address in future work.

\section{Related Work}
While a wide variety of works are labeled ``novel view synthesis'', the exact tasks they are designed to solve can have subtle but important differences.
We consider object-centric NVS techniques~\cite{park2017transformation, chan2023generative, gu2023nerfdiff, ye2024consistent, gao2024cat3d} to be a separate task, and instead focus on scene-level NVS, which poses different challenges such as more complex real-world scenes, multiple objects occlusions, and a high dynamic range of scene geometry.
We categorize the different NVS tasks by the following 4 axes:
(i) operating on static scenes~\cite{rombach2021geometry, barron2022mip} vs. dynamic videos~\cite{gao2021dynamic, zhao2024pgdvs};
(ii) having a single input view~\cite{tucker2020single, gu2024control3diff} or multiple views~\cite{wu2024reconfusion, deng2022depth};
(iii) outputting a single novel view~\cite{seo2024genwarp, yu2023long} or multiple consistent views~\cite{yu2024polyoculus, muller2024multidiff};
(iv) the viewpoint difference between inputs and outputs can be short (most of the image content is shared)~\cite{wiles2020synsin, barron2022mip, szymanowicz2024flash3d}, medium (some of the image content is shared)~\cite{rombach2021geometry, seo2024genwarp}, or long (novel view contents are mostly unseen)~\cite{poseguideddiffusion, yu2023long}.

The distinction among these tasks is crucial for determining the choice of training data, scene representation, and model architecture.
In this work, we focus on static scenes with a single input view and a single novel view in the mid- or long- range.
Under this setting, NeRFs~\cite{mildenhall2021nerf} and 3DGS~\cite{kerbl3Dgaussians} become less common as they struggle to extrapolate from a single viewpoint, even when combined with generative models~\cite{tewari2023diffusion, szymanowicz2024flash3d}.
Instead, several works~\cite{wiles2020synsin, shih20203d, liu2021infinite, chung2023luciddreamer} use monocular depth estimation (MDE), warp the pixels into the target view, and use an inpainting model to fill in missing details.
This approach has several drawbacks~\cite{seo2024genwarp}, such as high sensitivity to depth estimation errors and loss of semantic image details.
Generative models circumvent these issues by implicitly modeling the correlations among viewpoints. Thus, they are increasingly becoming a vital element of state-of-the-art approaches.
GeoGPT~\cite{rombach2021geometry} was among the first NVS methods to use generative modeling: they apply an autoregressive transformer to sample novel views conditioned on a single input view. 

\begin{figure}
    \centering
    \includegraphics[width=0.92\linewidth]{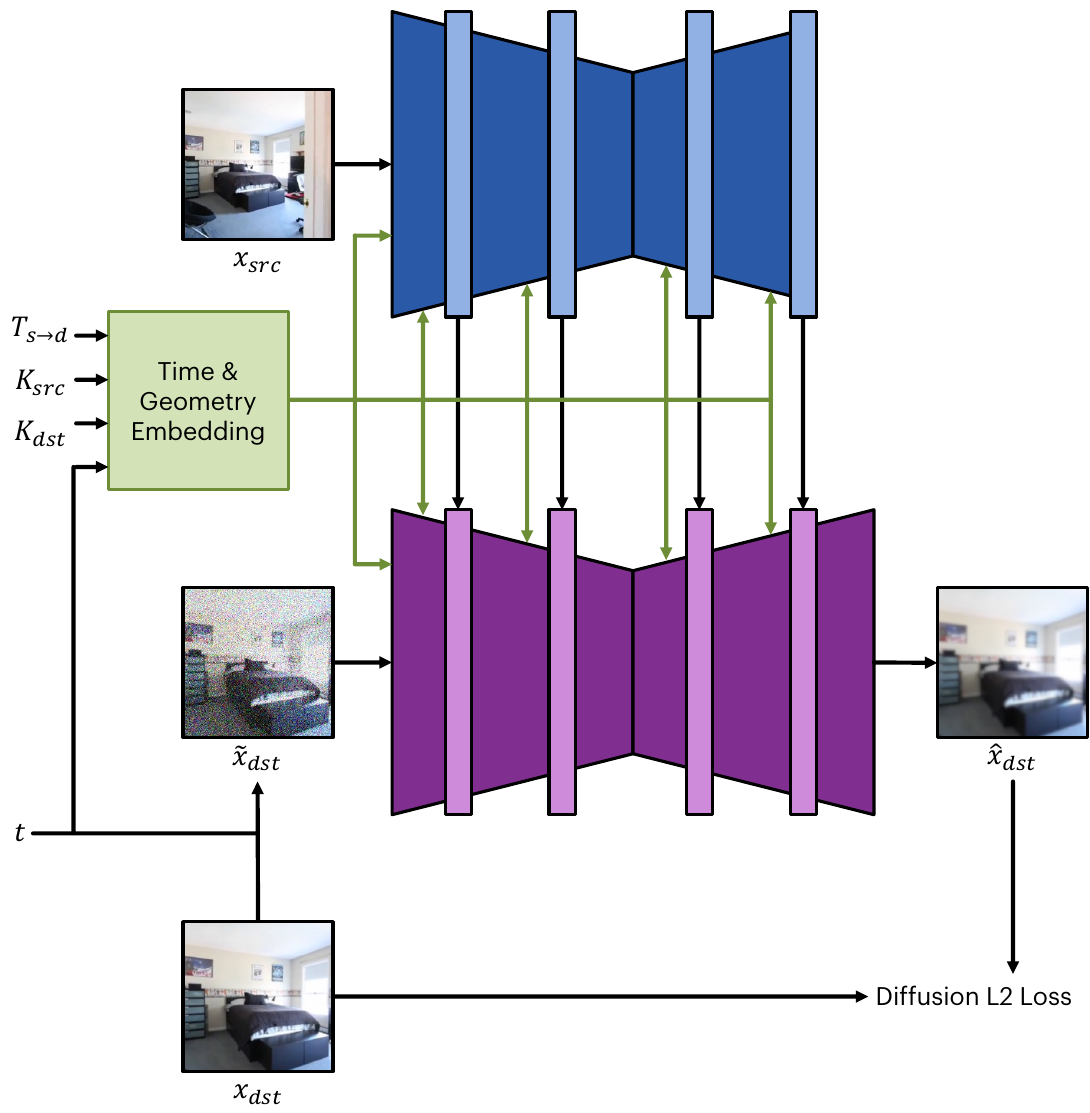}
    \caption{Overview of our model. The decoder (purple) learns to denoise the target view, using information from the source view provided by the encoder (blue) through cross-attention. Both models are aware of the diffusion timestep and scene geometry (green).}
    \label{fig:method}
\end{figure}

More recently, with the advent of powerful diffusion models~\cite{ho2020denoising, rombach2022high, karras2024analyzing}, many techniques have incorporated them into NVS pipelines.
In Pose-Guided Diffusion Models~\cite{poseguideddiffusion}, the authors utilize a pre-trained MDE~\cite{midas} to extract features from the input source view, which are then fed via cross-attention layers into a diffusion model that generates the target view. The cross-attention layers are constrained to pass information only along the epipolar lines, defined by the requested target viewpoint pose. 
Photometric-NVS~\cite{yu2023long} proposes a latent diffusion model with a two-stream architecture: two identical networks with shared weights process the source view and the noisy target views, exchanging information through cross-attention layers with pose information acting as query tokens.
GenWarp~\cite{seo2024genwarp} fine-tunes two copies of a pre-trained latent text-to-image diffusion model~\cite{rombach2022high}: one for encoding the source view, and another for generating the target. They use a CLIP~\cite{radford2021learning} image embedding of the source view instead of the text embedding, and further augment the network inputs with 2D coordinate embeddings, warped using MDE to match the source view coordinates.

\section{NVS Diffusion Model}
\label{sec:method}

\begin{figure}
    \centering
    \includegraphics[width=0.75\linewidth]{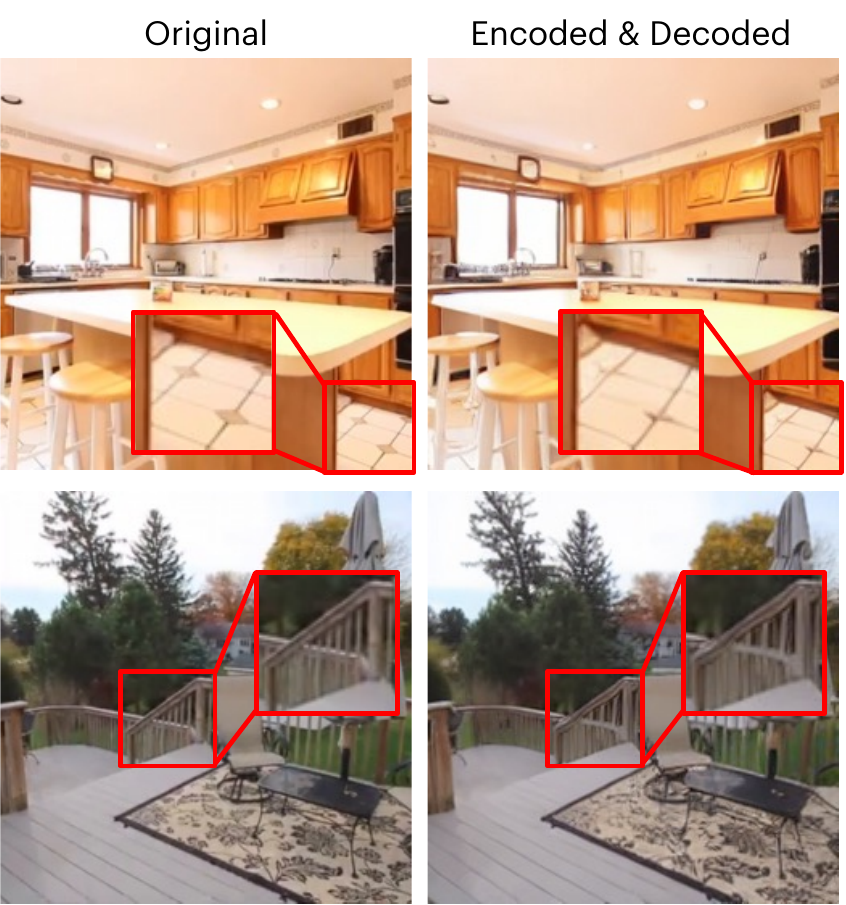}
    \caption{$256\times256$-pixel images from RealEstate10K~\cite{realestate10k}, encoded and decoded using the autoencoder from Stable Diffusion v1.4~\cite{rombach2022high}. Some areas with severe loss of detail are highlighted.} 
    \label{fig:sd-latent}
\end{figure}

We formalize the NVS task as follows: for a given source image ${\rvx_{src} \in \mathbb{R}^{C \times H \times W}}$, a camera transformation (extrinsics matrix) ${\mT_{s\to d} \in \mathbb{R}^{3 \times 4}}$, and camera intrinsic matrices ${\mK_{src}, \mK_{dst} \in \mathbb{R}^{3 \times 3}}$, our algorithm should generate novel view samples $\rvx_{dst} \sim p\left(\rvx_{dst}\;\middle|\;\rvx_{src}, \mT_{s\to d}, \mK_{src}, \mK_{dst}\right)$.
Most multi-view datasets contain extrinsics matrices $\mT_{src}, \mT_{dst}$ for each view.
These matrices transform global coordinates to the view-specific camera coordinates.
The transformation matrix between them can be obtained by $\mT_{s\to d} = \mT_{dst} \mT_{src}^{-1}$.\footnote{The inversion is done for a canonical $4\times4$ matrix representation.}

\subsection{Base Architecture}
\label{sec:arch}
Diffusion models~\cite{ho2020denoising, song2019generative} have emerged as a powerful tool to sample images from intricate conditional distributions such as text-to-image generation~\cite{rombach2022high, saharia2022photorealistic} and image restoration~\cite{kawar2022denoising, zhu2023denoising}.
This makes them an ideal candidate for NVS, because they enable powerful sampling with a relatively simple training objective~\cite{vincent2011connection, sohl2015deep, ho2020denoising}.
Diffusion model training requires a paired dataset of conditions ($\rvx_{src}, \mT_{s\to d}, \mK_{src}, \mK_{dst}$) and samples ($\rvx_{dst}$), adding Gaussian noise to the samples, and training a network to remove the noise given the conditioning signals.

We propose using two parallel U-Net~\cite{ronneberger2015unet} architectures with attention layers~\cite{vaswani2017attention}: an encoder to process $\rvx_{src}$, and a decoder to generate $\rvx_{dst}$.
The encoder acts as a feature extractor, while the decoder denoises the destination image $\rvx_{dst}$ as in common diffusion pipelines.
To condition the denoising process on the source image, we expand the self attention layers in the decoder to perform joint self and cross-view attention: query tokens from $\rvx_{dst}$ attend to key and value tokens from both $\rvx_{dst}$ itself and 
from $\rvx_{src}$ as extracted by the encoder.
We train both U-Nets jointly, leading the encoder to best provide features for NVS in an end-to-end manner.
Features are taken from the encoder at multiple layers and resolutions, enabling the transfer of global semantic information, as well as high resolution details.
We choose not to share weights between the two U-Nets~\cite{yu2023long}, nor build upon pre-trained models~\cite{poseguideddiffusion, seo2024genwarp} (\textit{e.g.}, text-to-image models or depth estimators).
This allows us to focus on end-to-end NVS without retaining artifacts from different objectives which may hinder NVS performance.

Additionally, we opt to perform the NVS diffusion process in the pixel space, and not in a latent space~\cite{seo2024genwarp} of a pre-trained autoencoder model~\cite{rombach2022high}.
We believe this to be an essential prerequisite for accurate correspondence to source image details, since simply encoding an image and then decoding it leads to a significant loss of detail. We show an example of this in \autoref{fig:sd-latent}.
Similar observations about autoencoder latent spaces were also made in different contexts~\cite{wang2023towards}.
As an alternative to the speedup that latent space compression provides, we use a cascaded diffusion design~\cite{ho2022cascaded}: we train a base model that receives $\rvx_{src}^{LR}$ at a low resolution and generates a novel view $\rvx_{dst}^{LR}$,
and a super-resolution (SR) model that receives $\rvx_{dst}^{LR}$ (with noise conditioning augmentation~\cite{ho2022cascaded}), and generates a high-resolution version of it $\rvx_{dst}^{HR}$.
The SR model has the same architecture as the base model, differing only by having a smaller number of channels, and receiving the high-resolution source $\rvx_{src}^{HR}$ as an additional input, allowing it to retain high-frequency details from the source view.
We provide a depiction of our base model architecture in \autoref{fig:method}, and of our SR model and end-to-end system in \autoref{app:implementation}.

\begin{table}
  \centering
  \begin{tabular}{@{}lcccc@{}}
    \toprule
    Geometry Encoding & FID ($\downarrow$) & PSNR ($\uparrow$) \\
    \midrule
    None & $5.75 \pm 0.04$ & $13.39 \pm 0.02$ \\
    Epipolar & $4.14 \pm 0.06$ & $17.43 \pm 0.03$ \\
    Pose & $3.00 \pm 0.02$ & $21.11 \pm 0.04$ \\
    Pose + Epipolar & $2.87 \pm 0.04$ & $21.15 \pm 0.04$ \\
    Pose + Depth & $2.99 \pm 0.01$ & $21.29 \pm 0.04$ \\ 
    Pose + Coordinate & $3.08 \pm 0.04$ & $21.06 \pm 0.04$ \\
    \bottomrule
  \end{tabular}
  \caption{Geometry encoding ablation. Metrics were computed $5$ times for randomly sampled $10$K source-target pairs.}
  \label{tab:geom}
\end{table}

We use the UNet architecture proposed in EDM2~\cite{karras2024analyzing}, due to its impressive performance and training stability.
In addition to the attention expansion and some hyperparameter choices (detailed in \autoref{app:implementation}), we perform two additional changes to the architecture:
(i) we add a single attention layer at the second-highest resolution to enhance fine detail fidelity;
and (ii) we encode the geometric conditioning information ($\mT_{s\to d}, \mK_{src}, \mK_{dst}$) into the network in several ways, as described in the next section.

\subsection{Geometry Encoding Ablation}

\begin{figure}
    \centering
    \includegraphics[width=0.8\linewidth]{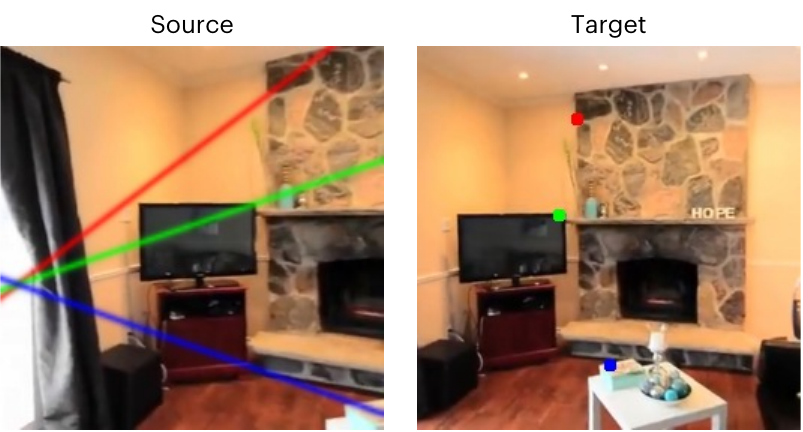}
    \caption{We select 3 random points in the target view, and show their epipolar lines in the source view in the corresponding color. In epipolar attention, we add a cross-attention bias relative to the proximity of the source view token to the target epipolar line.}
    \label{fig:epipolar}
\end{figure}

GeoGPT~\cite{rombach2021geometry} demonstrated a reasonably good end-to-end generative NVS autoregressive model, using a simple encoding of the camera pose as model input.
They make the claim that there is no need for further geometry-specific modules such as depth estimation or geometry-aware feature-matching layers.
Despite these findings, many recent diffusion-based NVS methods~\cite{yu2023long, poseguideddiffusion, seo2024genwarp} reach superior results by incorporating advanced forms of geometric bias, questioning whether this claim still holds.
Here, we conduct thorough ablations on four types of 3D geometry encoding methods, and assess their impact.
We train all model variants on RealEstate10K~\cite{realestate10k}, and provide more details in \autoref{app:geometric}.

\paragraph{Pose Embedding.} This is the simplest form of geometry encoding, often used in conjunction with other methods~\cite{rombach2021geometry, yu2023long}.
We simply take the scalars of the camera pose matrices $\mT_{s\to d}, \mK_{src}, \mK_{dst}$, and encode them into a shallow perceptron that shifts the model's activations, similar to diffusion timestep embedding~\cite{ho2020denoising, karras2022elucidating, karras2024analyzing}.
In our work, we also normalize these scalars by their dataset-wide mean and standard deviation before embedding them.

\paragraph{Epipolar Attention.} In the cross-attention layers, target view tokens attend to their source view counterparts based on their visual features.
With epipolar attention, we attempt to enrich the source-target correlation with camera pose information by modifying the cross-attention maps.
For each position in the target view, we can use the camera pose information to find the relevant source view positions along its epipolar line, guiding the model towards features that constitute geometrically correct information.
However, these correlations may amplify irrelevant information, as only a handful of points on the epipolar line will correspond to the query's target view position.
Thus, we prefer the epiopolar correlations to act as a learnable bias for the attention matrix, enabling the model to select where the epipolar information is useful during training.
Specifically, we use an implementation similar to~\cite{poseguideddiffusion}, with two small but critical modifications:
First, our epipolar correlations act as an additive bias to the attention matrix, instead of a multiplicative transformation.
This does not zero out correlations outside the epipolar line, which may have critical semantic significance.
Second, each attention head has its own independently learned mixing parameters.
This enables the model to produce both geometrically dependent features, and semantically significant features, based on the parameter values for each attention head.
We illustrate this method in \autoref{fig:epipolar}.

\begin{figure}
    \centering
    \includegraphics[width=0.8\linewidth]{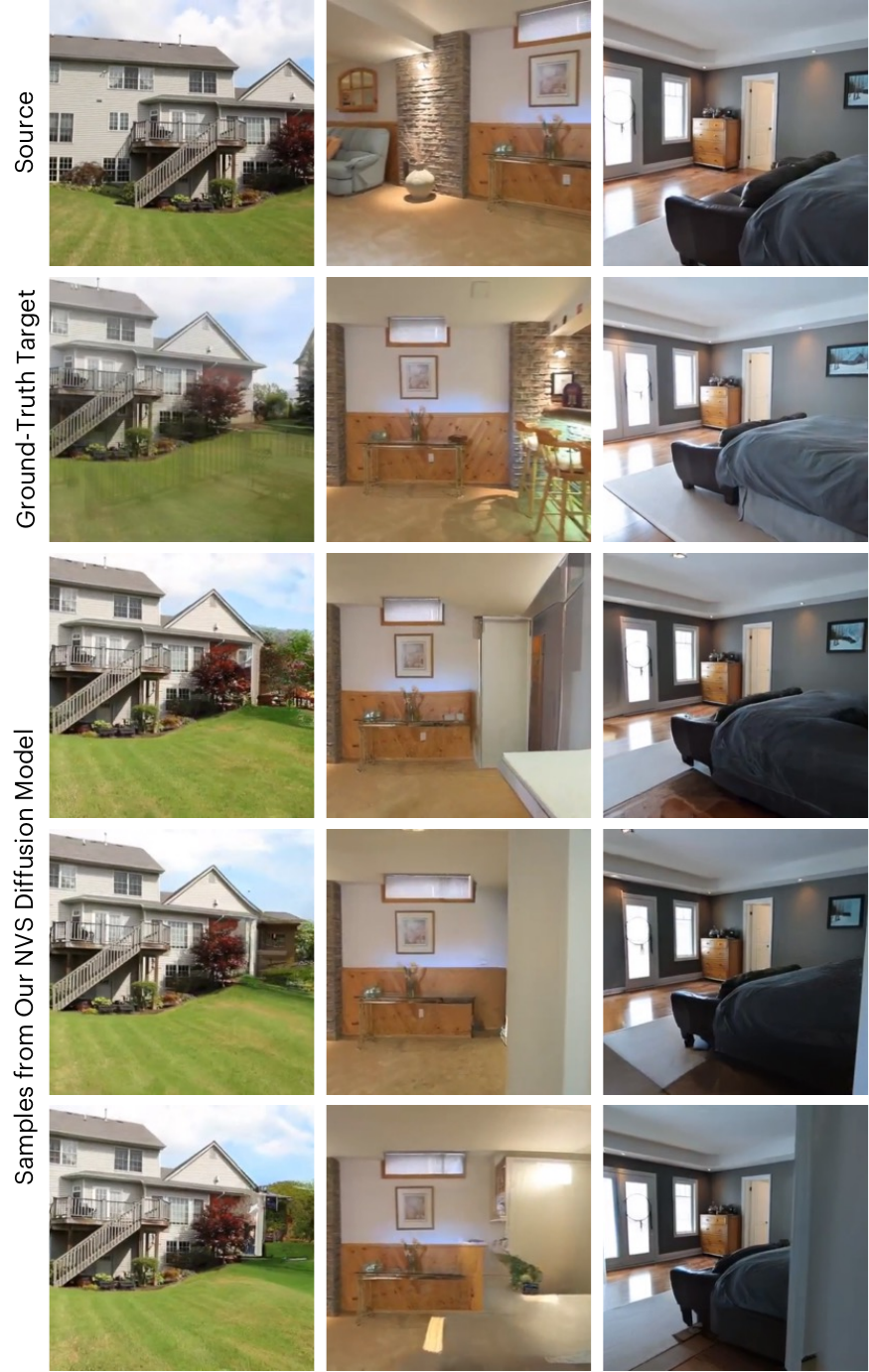}
    \caption{Different NVS results for the same input sampled from our model. Details from the source view are kept in all samples, and diverse realistic options are generated in newly visible areas.
    }
    \label{fig:multiple}
\end{figure}

\paragraph{Monocular Depth Estimation.} MDE is used in many key NVS works~\cite{rombach2021geometry, seo2024genwarp}, either as an additional input, or for warping intermediate predictions and network features.
However, we believe the information contained in a monocular depth prediction can be learned internally within the NVS model, given enough model capacity, training data, and time.
We conduct experiments using a DepthAnythingV2~\cite{depth_anything_v2}, a recent state-of-the-art MDE model, attaching its prediction on the source view as additional input to our encoder.

\paragraph{Coordinate Warping.} MDE can also be used to warp coordinate embeddings, creating additional inputs for both the encoder and decoder, which can assist in feature-matching. This technique was proposed in~\cite{seo2024genwarp}, and we test it using the DepthAnythingV2 depth prediction~\cite{depth_anything_v2}.

\begin{figure*}
    \centering
    \includegraphics[width=\linewidth]{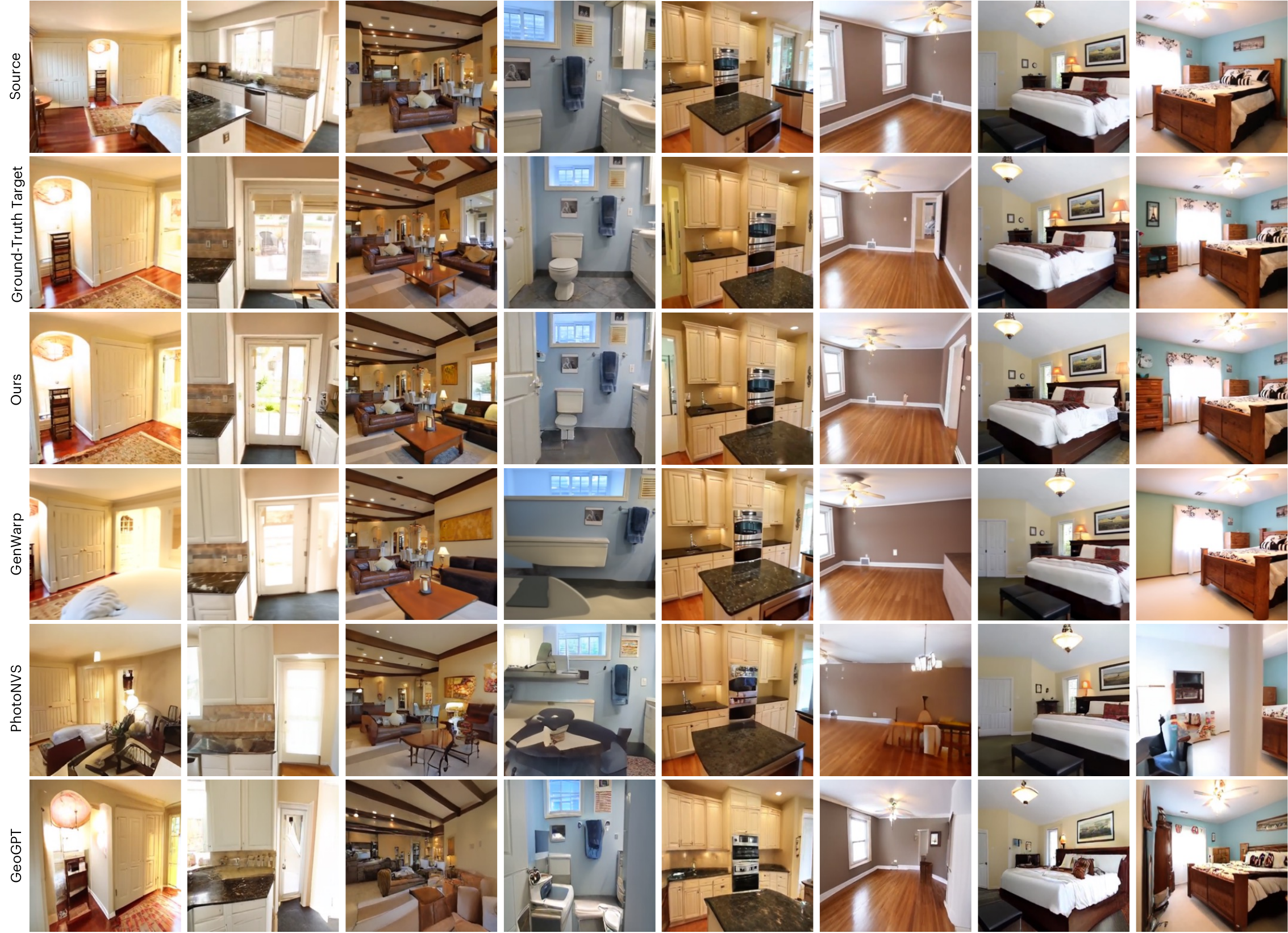}
    \caption{Comparison between our method and previous state-of-the-art approaches in NVS on RealEstate10K~\cite{realestate10k}.
    }
    \label{fig:comp}
\end{figure*}

\paragraph{Results.} In \autoref{tab:geom}, we compare the various geometry encoding methods on our base diffusion model, handling $64 \times 64$-pixel images, uniformly sampled from the RealEstate10K~\cite{realestate10k} test set.
We start with a geometrically-uninformed baseline (no use of $\mT_{s\to d}, \mK_{src}, \mK_{dst}$), and then add pose embedding and epipolar attention to it.
While both methods help, pose embedding produces far better results. 
Thus, we utilize pose embedding in conjunction with each of the other methods.
While they mostly improve FID~\cite{fid} and PSNR, the improvement is often small and not noticeable in qualitative evaluations.
As a result, we prefer the pose embedding technique for its simplicity and effectiveness, and use this variant in the following sections.

\subsection{Comparison to Previous Methods}
\label{sec:comparison}
In our ablations, we train the EDM2~\cite{karras2024analyzing} architecture with a limited batch size and number of iterations, and without using exponential moving average (EMA).
In this section, we scale up our model (with the pose embedding technique): we double the batch size and quadruple the number of training iterations, use EDM2's Power function EMA~\cite{karras2024analyzing}, and apply classifier-free guidance (CFG)~\cite{ho2021classifier} using a separate unconditional denoiser. We provide more comprehensive details and hyperparameter choices in \autoref{app:implementation}.

\begin{table}
  \centering
  \begin{tabular}{@{}lccccc@{}}
    \toprule
    &  \multicolumn{2}{c}{Mid-range} & \multicolumn{2}{c}{Long-range} \\
    Method & FID $\downarrow$ & PSNR $\uparrow$ 
    & FID $\downarrow$ & PSNR $\uparrow$ \\
    
    \midrule
    GeoGPT~\cite{rombach2021geometry} & $6.43$ & $14.06$ 
    & $7.22$ & $13.13$
    \\
    PhotoNVS~\cite{yu2023long} & $7.12$ & $13.32$ 
     & $9.22$ & $12.05$  \\
    GenWarp~\cite{seo2024genwarp} & $5.91$ & $13.43$
    & $7.38$ & $12.10$ 
    \\
    
    VIVID (Ours) & $\textbf{2.89}$ & $\textbf{17.36}$ & $\textbf{3.89}$ & $\textbf{15.21}$ \\
    \midrule
    Source View & $2.58$ & $13.12$ 
    & $3.00$ & $11.91$ 
    \\
    \bottomrule
  \end{tabular}
  \caption{Comparison to previous methods. Evaluation is done on ${10}$K source-target pairs from RealEstate10K~\cite{realestate10k}.}
  \label{tab:comp}
\end{table}

We compare our results in terms of image quality using FID~\cite{fid} and distortion from the ground truth using PSNR.
For a thorough evaluation, we generate $10$K novel views from each tested method based on randomly sampled source images from the RealEstate10K~\cite{realestate10k} test set.
To sample a ground-truth novel view and its corresponding camera transformation, two distinct ranges are identified, following GenWarp~\cite{seo2024genwarp}:
mid-range and long-range, corresponding to target views that are $30$-$60$ and $60$-$120$ frames away from the source, respectively.
We use our base and SR models to generate $256\times256$ images.
We compare FID and PSNR of images generated from our model with the results of state-of-the-art generative NVS methods: GenWarp~\cite{seo2024genwarp}, Photometric-NVS~\cite{yu2023long}, and GeoGPT~\cite{rombach2021geometry}.
We additionally report the FID and PSNR of simply outputting the source view for any requested transformation as a naïve frame of reference. We expect the source view to achieve high perceptual quality, as the source and destination images are sampled from the same distribution, yet have low PSNR.
We present the results in \autoref{tab:comp}, showing a substantial improvement of our method (more than $24\%$) over the previous state-of-the-art across all ranges and metrics. 
We show qualitative results in \autoref{fig:results} and \autoref{fig:comp}, and provide additional metrics (\textit{e.g.}, joint FID~\cite{devries2019evaluation}), details, and discussion in \autoref{app:comparison}.

Furthermore, since our approach uses a probabilistic diffusion model, it can sample from the distribution of novel views given a source view and a camera transformation. This distribution can have significant variance in its results, especially for large camera transformations.
In \autoref{fig:multiple}, we demonstrate our model's ability to estimate this distribution by drawing multiple samples for the same input.

Our findings demonstrate that applying the diffusion process in the pixel space (rather than a latent space) is crucial for NVS.
Using latent diffusion causes a loss of fine details in the source image that is not recoverable, as demonstrated in \autoref{fig:sd-latent}, and validated in previous methods' results in \autoref{fig:comp} (as they use latent diffusion).
Additionally, we hypothesize that the use of a dedicated encoder, implicitly trained to encode relevant NVS information, is a major advantage of our method over alternatives such as Photometric-NVS~\cite{yu2023long}.

\section{NVS Training on Single-View Data}
\label{sec:generalization}
We have shown impressive NVS results for a model trained on the RealEstate10K~\cite{realestate10k} training set, and tested on its corresponding test set, following the standards set in previous literature.
However, towards the goal of achieving a general-purpose NVS model, these training and evaluation strategies suffer from a few key shortcomings.

First, both training and evaluation data consist almost entirely of scenes inside houses. This is a biased and limited class of scenes, and does not cover the general real-world image distribution.
Other large-scale multi-view datasets~\cite{liu2021infinite, yu2023mvimgnet} similarly suffer from limited diversity of both semantic contents and camera trajectories.
This is understandable, since capturing multiple views of static scenes at a large scale is challenging, especially compared to the quantities achieved in single-view image datasets~\cite{kuznetsova2020open, schuhmann2022laion}.

\begin{figure}
    \centering
    \includegraphics[width=0.95\linewidth]{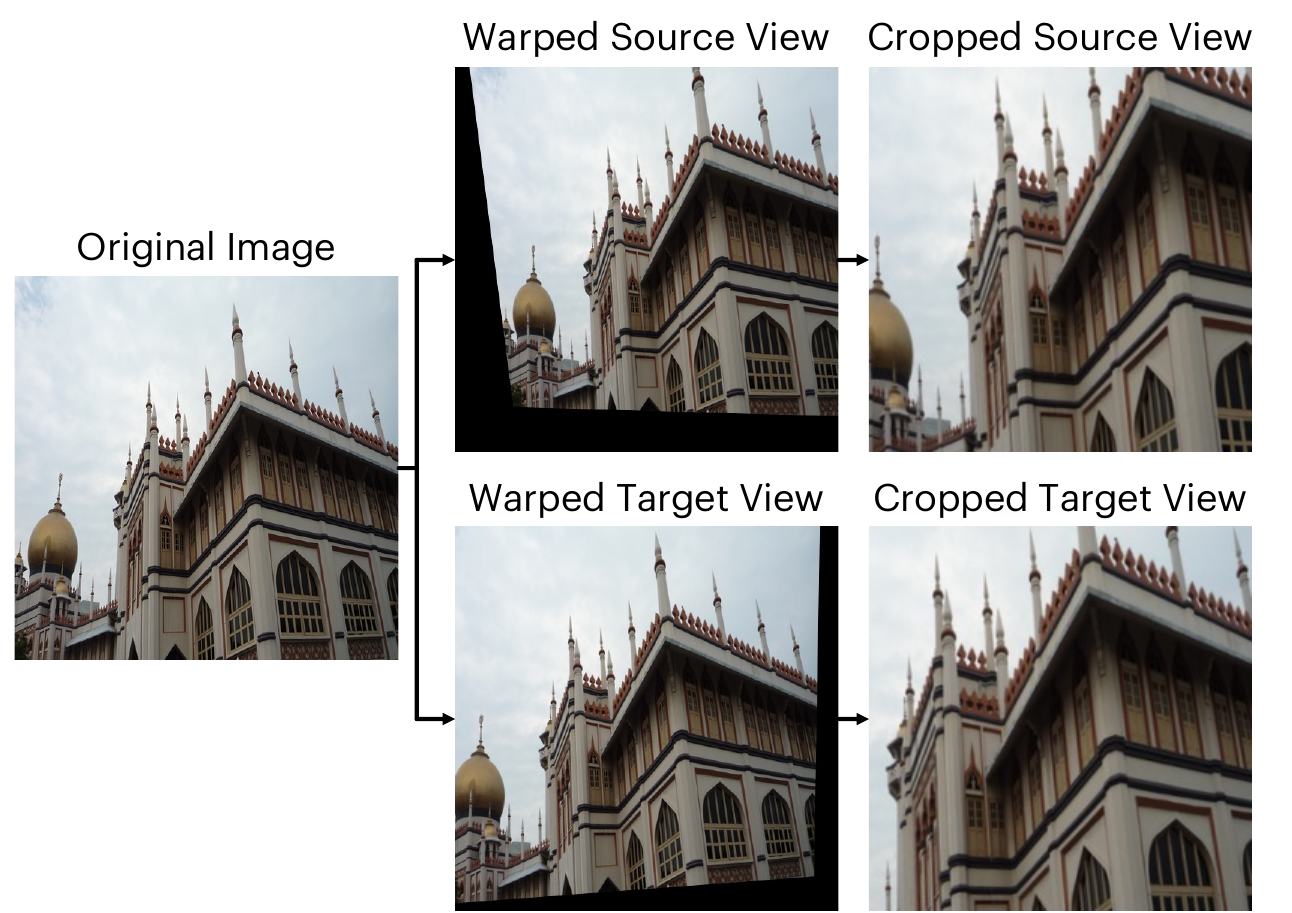}
    \caption{Example of our proposed homography augmentation. We randomly sample two camera rotation matrices, apply the transforms, and then perform the same crop to both views. In this example, the camera rotates to the right and slightly up from the source to the target view.}
    \label{fig:homography}
\end{figure}

Second, the multiple views in a given scene originate from frames inside a ${360 \times 640}$-pixel video touring a house.
The tour videos predominantly capture empty houses with no movement, which is a good feature as different frames in the video can simulate multiple cameras capturing the same static scene.
However, the data is low in resolution, and it also undergo significant video compression that further degrades the quality of the model input and supervision.

Third, the camera positions in the different frames are not measured during video capture.
Existing multi-view datasets offer camera extrinsics matrices which are synthetically generated through Structure-from-Motion (SfM) algorithms.
While SfM algorithms often produce impressive results, they have their own margin of error and failure cases.

Lastly, even if we ignore their inaccuracies, SfM algorithms have another fundamental limitation: scale ambiguity -- they cannot accurately recover the metric scale of the entire scene.
Furthermore, because we focus on NVS with a single image input, the NVS model will also be limited, as it cannot accurately infer the scale of its input view in relation to the scale of the camera transformation matrix.
RealEstate10K~\cite{realestate10k} and others~\cite{sargent2024zeronvs, seo2024genwarp} have addressed this problem by normalizing the scales of the translation vectors inside the camera matrices, either at the scene level or at the single image level.
These methods stabilize the NVS model behavior and measurably improve its results, but they are still based on heuristic techniques and are prone to error, resulting in mismatches between the camera transformation $\mT_{s\to d}$ and the ground-truth target view we compare against.
These errors and mismatches have an increasing effect when considering data from multiple sources, cameras, dynamic ranges, lighting conditions, and SfM algorithms.

To mitigate some of these issues, we propose a novel NVS training scheme that makes use of both multi-view and single-view data.
We apply it to our model, and demonstrate its benefits on out-of-domain test data.

\begin{table}
  \centering
  \begin{tabular}{@{}lcccc@{}}
    \toprule
    Percentage \quad\quad\quad & FID ($\downarrow$) & PSNR ($\uparrow$) \\
    \midrule
    $0\%$ & $36.14$ & $ 14.38$ \\
    $10\%$ & $\mathbf{31.98}$ & $\mathbf{14.52}$ \\
    $25\%$ & $35.27$ & $14.45$ \\
    \bottomrule
  \end{tabular}
  \caption{NVS performance on $20$ out-of-domain scenes, with varying percentages of training data stemming from single-view images with our proposed augmentation.}
  \label{tab:mipnerf}
\end{table}

\subsection{Single-View Augmentation}
Compared to multi-view data, high-resolution single-view images are abundantly available.
Given a single image, we would like to simulate multiple views of a scene, beyond the trivial identity transform.
To do so, one might consider taking smaller crops of high-resolution images.
However, note that different overlapping crops do not represent any geometrically grounded camera transform.
Camera translations can result in dis-occluded areas depending on the depth difference of different objects in the scene, with contents that are unseen in the original image.
Therefore, we focus on simulating camera rotations.

To that end, we apply a homography transform on a high-resolution image, which warps the image's contents according to a randomly sampled camera rotation matrix.
We apply the transform twice, for two randomly chosen rotation matrices, generating a source view and a target view.
This produces warped images with undefined areas near the edges, as they should contain contents that were not visible in the original view.
However, for a crop far enough from image boundaries, and small enough camera rotations, the resulting view is contained entirely in the original image.
We then compare this to the corresponding crop in the target image, successfully simulating a camera rotation.
We provide an illustrative example of this homography transform in \autoref{fig:homography}.

\begin{figure}
    \centering
    \includegraphics[width=0.95\linewidth]{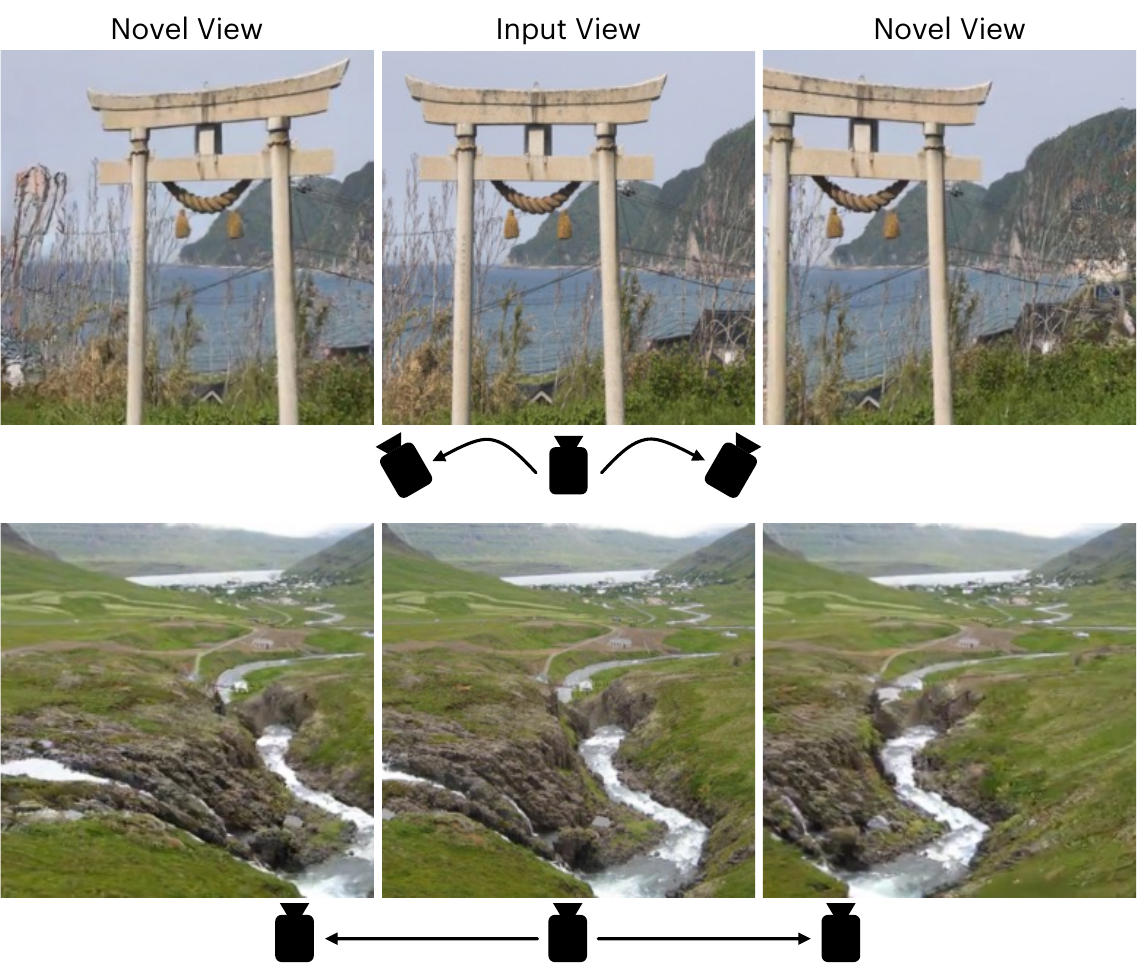}
    \caption{Our NVS model performs camera rotation (top) and translation (bottom) on OpenImages, a single-view dataset.}
    \label{fig:generalization}
\end{figure}

With this data augmentation strategy and a single-view image dataset, we produce a large amount of source-target view pairs with diverse and accurately-labeled camera rotations.
We then train our NVS model on a mixture of multi-view and single-view datasets, attempting to capture the benefits of both: real-world camera movement with ground-truth views from multi-view datasets, and rich diverse image contents with non-trivial camera transforms from single-view datasets.
While the fact that homography transforms can simulate camera rotations has long been known, to the best of our knowledge, this is the first work to utilize it as part of a NVS model training scheme.

\subsection{Experimental Results}
We enrich our original model training scheme presented in \autoref{sec:method} with the use of single-view data with our proposed augmentation.
We choose the OpenImages v5 dataset~\cite{kuznetsova2020open} as our source of single-view images, owing to its diversity and large size (more than $9$ million images), and continue with RealEstate10K~\cite{realestate10k} as the source of multi-view images.
To avoid our model collapsing to only learning rotations, we need to strike a balance between the two datasets.
To that end, we retrain our base model with a percentage of the training data coming from single-view OpenImages scenes: $0\%$ (original, RealEstate10K only), $10\%$, and $25\%$.
We provide more training details in \autoref{app:implementation}.

Then, to evaluate the generalization capabilities of our NVS model, we test it with input images and camera transformations that are out-of-domain with respect to RealEstate10K.
We perform qualitative evaluation with input source views from the OpenImages test set and camera transformations that include both rotation and translation, and
show results in \autoref{fig:generalization} (using the $10\%$ model).

Moreover, to obtain quantitative NVS performance metrics on out-of-domain data, we draw source-target view pairs from 20 scenes coming from 3 multi-view datasets (LLFF~\cite{mildenhall2019llff}, MipNeRF-360~\cite{barron2022mip}, and Ref-NeRF~\cite{verbin2022refnerf}), and evaluate FID and PSNR for each of our models.
As evident \autoref{tab:mipnerf}, the $10\%$ model exhibits a significant improvement over the $0\%$ one in both FID and PSNR. 
This improvement shows that the model manages to benefit from the diverse single-view training data, while remaining true to the NVS framework due to our proposed augmentation.
However, increasing the percentage to $25\%$ leads to degraded performance, possibly due to an over-representation of rotation transforms in the data.

\section{Conclusion}
\label{sec:conclusion}
In this work, we explored and analyzed the many options proposed in the literature for designing an end-to-end novel view synthesis approach based on a generative model.
We utilized a leading diffusion model architecture~\cite{karras2024analyzing} as our backbone network, adapted it for the NVS task, and ablated on options for encoding the geometric information input.
Our experiments show that while some sophisticated options can slightly enhance performance, the simple scalar embedding option works very well and the differences are mostly negligible.
In fact, we show that our generative modeling choices such as utilizing a powerful architecture, passing information across views through attention, and operating in the pixel space, substantially upgrade the quality of NVS results compared to previous state-of-the-art methods.

Our approach achieves SOTA results in the widely accepted RealEstate10K~\cite{realestate10k} benchmark.
Nevertheless, we highlight the shortcomings and potential pitfalls of this benchmark towards a comprehensive general-content NVS approach.
We address some of these shortcomings with a novel training scheme enabling NVS model training on the abundantly available single-view image datasets, benefiting from their content diversity while maintaining the geometric capabilities learned from the multi-view datasets.
We demonstrate that our proposed scheme can significantly enhance NVS performance on out-of-domain data.

In future work, we anticipate the continued improvement of NVS capabilities through the utilization of more modern and advanced generative models.
We also hope that our method's limitations, which are also common in NVS literature, can be addressed.
These limitations include generalization to camera trajectories that are uncommon or nonexistent in the training set, as our proposed augmentation does not introduce any camera translation.
In addition, our method is inherently dependent on  the scene scales provided in its training set, which could be problematic if they are obtained using Structure-from-Motion (SFM) techniques without some form of normalization~\cite{realestate10k, sargent2024zeronvs}.

{
    \small
    \bibliographystyle{ieeenat_fullname}
    \bibliography{main}
}

\clearpage
\setcounter{page}{1}
\appendix
\maketitlesupplementary

\section{Implementation Details}
\label{app:implementation}

\subsection{Model Architecture \& Training}
\label{app:hyperparams}

\begin{table*}
  \centering
  \begin{tabular}{@{}lccccc@{}}
    \toprule
    Model & Batch Size & Training Iterations & Base Channel Width & Learning Rate & EMA\\
    \midrule
    Base (Ablation) & $512$ & $2^{18}$ & $128$ & $0.012$ & - \\
    Base (Final) & $1024$ & $2^{20}$ & $128$ & $0.012$ & $0.1$ \\
    Base (For CFG) & $1024$ & $2^{19}$ & $128$ & $0.012$ & $0.05$ \\
    SR & $256$ & $2^{20}$ & $64$ & $0.01$ & $0.05$ \\
    
    \bottomrule
  \end{tabular}
  \caption{Training Hyperparameters}
  \vspace*{-0.5em}
  \label{tab:arch}
\end{table*}

Our architecture is based on the U-Net~\cite{ronneberger2015unet} outlined in EDM2~\cite{karras2024analyzing}. To facilitate the processing of both the source image and the noisy destination, we duplicate our U-Net architecture into two dedicated networks, an encoder and a decoder, as outlined in \autoref{sec:arch} and ~\autoref{fig:method}. Specifically, we employ joint-attention between the encoder's deep features and the decoder's, in which keys and values are computed for both the encoder and decoder stream and concatenated, whereas queries are computed only from the decoder.
We find empirically that joint-attention performs better than self-attention followed by cross-attention.
We use the default 3 layer per resolution, using resolutions $[64, 32, 16, 8]$ and $[256, 128, 64, 32]$ for the base and SR models respectively.
Beyond the existing attention layers, we add a single attention layer at the second-highest resolution ($32$) on the base model only to enhance fine detail fidelity transferred between encoder and decoder through the attention mechanism. 

We condition both the encoder and decoder networks on the diffusion timestep.
If time conditioning is omitted from the encoder network, it would only need to be run once for each generation, as the source image and geometry do not change.
In preliminary experiments we find that this change has a slight negative effect on the results, as expected due to the encoder's lack of ability to match its extracted features to the current diffusion timestep.
For simplicity, we focus on the conditioning both networks on the timestep, seeking higher quality results, and we leave the exploration of the more efficient option to future work.

We base our hyperparameter choices on EDM2~\cite{karras2024analyzing}, making no change across our experiments to: number of residual blocks, channel multiples, sampling algorithm, noise schedule, number of inference diffusion steps, and Adam optimizer hyperparameters.
We do not use dropout and use training noise level distribution with a log-normal distribution where mean and standard deviation are $P_{mean}=-0.8$ and $P_{std}=1.6$ respectively.
\autoref{tab:arch} lists our experiment-specific training hyperparameter choices. 

During training, we iterate over scenes in the RealEstate10K dataset, uniformly choosing two views as the source and destination views regardless of the distance between the frames.
During inference, we first uniformly sample the source view and later uniformly sample a destination view that is 30-60 (mid) or 60-120 (long) frames apart. Source views that do not have possible destinations at the appropriate frame distance are filtered out.
We have noticed that tailoring the distance between frames to 30-120 (the relevant distance) during training is not beneficial for image quality or metrics.

For sampling, we employ Classifier-Free Guidance~\cite{ho2021classifier} (CFG) using a separate unconditional diffusion model, following EDM2~\cite{karras2024analyzing}, instead of the more commonly used single model with label dropout. The unconditional model does not contain the encoder network, instead receiving zeros as pose encoding and joint-attention features. We choose a CFG coefficient of $1.5$ for generating samples from RealEstate10K, and $2.0$ for the experiments in \autoref{sec:generalization}. The EMA values are not tuned further for CFG.

\begin{figure}
    \centering
    \includegraphics[width=0.9\linewidth]{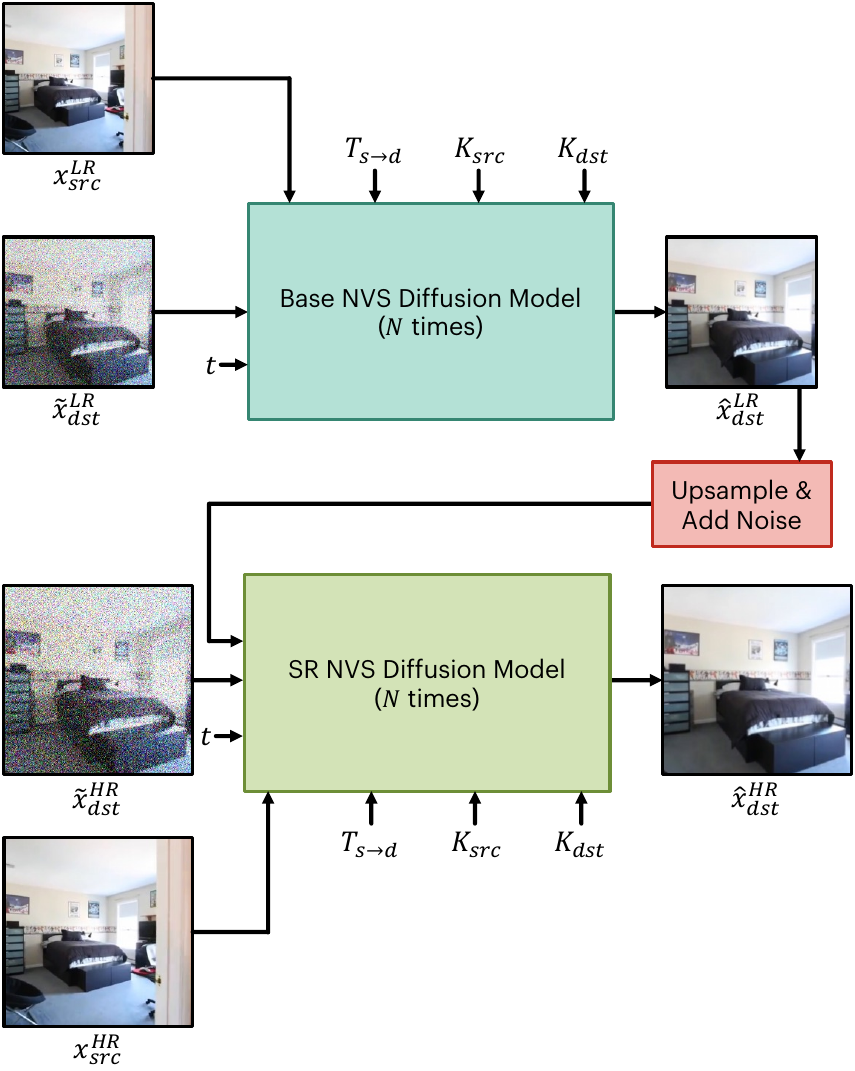}
    \caption{End-to-end inference using our base and super-resolution (SR) NVS diffusion models.}
    \label{fig:sr_method}
\end{figure}

\subsection{Cascaded Diffusion and Super-Resolution}
\label{app:sr}
As detailed in \autoref{sec:arch}, we use a cascaded diffusion model design to retain the pixel-space information used in NVS.
Thus, a super-resolution (SR) model is trained separately from the base model. 
Specifically, the task of NVS super-resolution is a relatively simple task, having both the low frequency information from the low-resolution conditioning image and the high-frequency semantics and texture from the source view.
For this reason, we opt to use an SR model with smaller capacity and batch size, and a shorter training time.
Additionally, to alleviate the distribution gap between ground-truth and generated low-resolution conditions, we add white Gaussian noise to the conditioning image before using it as an input to the SR model, as proposed in~\cite{ho2022cascaded}. We find that noise of standard deviation $0.25$ works well.
Our end-to-end inference system (including the base and SR models) is depicted in \autoref{fig:sr_method}.

\section{Geometry Encoding Ablation Details}
\label{app:geometric}

\subsection{Pose Embedding}
To apply our pose embedding, we encode the camera information into a $20$-element encoding, as such: 
\begin{equation}
    E_{pose} = \texttt{cat}\left(\texttt{flatten}\left(\mT_{d
    \to s}\right), f_{src}, p_{src}, f_{dst}, p_{dst}\right)
\end{equation}
Where $f$ and $p$ are the focal length and principle point extracted from the intrinsic matrixes $\mK_{src}$ and $\mK_{dst}$.
The embedding conditions each U-Net block using a single fully-connected layer and summation, similarly to class embedding conditioning in EDM2~\cite{karras2024analyzing}.
The embedding is normalized to zero mean and standard deviation of one using precomputed estimations based on $64\times64$ source-destination pairs from RealEstate10K.
For the $256\times256$ SR model the embedding is altered by extrapolating the statistics of $64\times64$ intrinsics to $256\times256$ intrinsic matrices. The extrinsics do not change for the different image resolutions.

\subsection{Epipolar Attention Bias}
We test the use of epipolar attention bias as a type of geometric encoding.
Our implementation is inspired by the epipolar attention used in~\citep{poseguideddiffusion}, in which an epipolar attention matrix is computed by using a soft cutoff (sigmoid) function on the epipolar distance between any two pixels in the source and destination images. In~\citep{poseguideddiffusion}, the epipolar attention matrix multiplies the cross-attention matrices in all layers, effectively zeroing out the feature correlations that do not fit the geometry.
We opt to use the epipolar attention matrix as an attention bias instead, replacing the multiplication operator with addition.
A soft attention bias enables the network to transfer information between the source and destination streams even when the information does not strictly fit the geometric composition, while multiplication by zeroes strictly prohibits that.
Furthermore, we allow each attention head to learn its own mixing parameter for the epipolar attention, enabling the network to learn separate heads for strictly geometric correspondence and semantically significant features. Specifically, we learn 4 scalars per head in each attention block, modulating the amplitude ($m$), temperature ($\tau$), cutoff ($c$), and bias ($b$) for mixing the epipolar attention bias. The following formula computes our epipolar attention bias ($\mA_{epipolar}$) using the epipolar distance matrix ($\mD_{epipolar}$) and the learnable mixing parameters:
\begin{equation}
    \mA_{epipolar} = m \cdot \sigma(\tau \cdot (c - \mD_{epipolar})) + b
\end{equation}
Specifically, the bias is needed due to the use of joint-attention, as the epipolar attention bias is only added to the ``cross'' part of the attention matrix (\textit{i.e.}, the attention between source and destination view features). The epipolar attention mixing parameters are initialized such that all elements of the epipolar attention bias are zero.

\subsection{Monocular Depth Estimation}
We use the small indoor metric model from DepthAnythingV2~\cite{depth_anything_v2} to estimate the source image depth. The depth-map is inverted, normalized by the maximum value and further normalized across the depth map distribution to zero mean and standard deviation of one, to have a similar magnitude to the input source image. While the depth normalization preserves the ordinality of the depth, its metric accuracy is effectively erased. In any case, since the depth prediction is not up to scale with the RealEstate10K poses, the metric accuracy of the MDE is not beneficial for NVS. 

\subsection{Depth Coordinate Warping}
We implement coordinate warping following the method outlined in GenWarp~\cite{seo2024genwarp}.
We start by creating a 2-dimensional grid for the source image's pixels.
We then warp (in 3 dimensions) this grid to the destination view using the monocular depth predicted with the same model used in the previous section.
As a result, we obtain a destination view grid where each pixel points to its matching coordinate from the source view grid.
Finally, we encode both the source and destination grids into $128$ Fourier feature maps, and concatenate them to the encoder and decoder inputs, respectively.
This increases the efficiency of the warping operation by skipping the interpolation step, instead relies on the network to learn the correlations.

\section{Single-Image Augmentation}
\label{app:augmentation}

\subsection{Augmentation Implementation Details}
We use the OpenImages v5 dataset~\cite{kuznetsova2020open} for the single-image augmentation experiment. We first center-crop and resize all images to $512\times512$. Then, we select the augmentation parameters for the source and destination views, creating two distinct camera poses and matching warping operations.
We uniformly sample the yaw, pitch, and roll\footnote{We use the COLMAP~\cite{schoenberger2016sfm, schoenberger2016mvs} coordinates, where $x$, $y$, and $z$ axes point left, down, and forward respectively.} as such:
(i) with probability $0.5$, we sample them from $[-5.5^{\circ}, 5.5^{\circ}], [-5.5^{\circ}, 5.5^{\circ}], [0^{\circ}, 0^{\circ}]$ and center crop to $384\times384$;
and (ii) otherwise, we sample them from $[-8.3^{\circ}, 8.3^{\circ}], [-8.3^{\circ}, 8.3^{\circ}], [-3.5^{\circ}, 3.5^{\circ}]$ and center crop to $320\times320$.
These parameters also determine the simulated geometry, by computing the rotation matrix corresponding to the sampled angles for both the source and destination views.
The intrinsic matrix is constructed to match the statistics of RealEstate10K images, where the focal length and principle point ate chosen as $(307.2, 307.2)$ and $(256, 256)$ for the $512\times512$ images.
Finally, the produced source and destination views are resize to the appropriate model resolution, and mixed into the training batches as a constant fraction of every training batch.

\subsection{Generalization Quantitative Analysis}
We use 20 scenes from the well-established datasets LLFF~\cite{mildenhall2019llff}, MipNeRF-360~\cite{barron2022mip}, and Ref-NeRF~\cite{verbin2022refnerf} to perform a qualitative evaluation of the benefit of our proposed single-image augmentation.
We use the preexisting camera poses and intrinsics, rescaled for each scene using a single scalar value. 
As a replacement for the ``mid'' and ``long'' distance between frames used in RealEstate10K, we divide the source-target pairs into ranges based on a certain LPIPS~\cite{zhang2018perceptual} threshold, chosen for each scene to mitigate the scene-level scale ambiguity issue.

\section{Method Comparison}
\label{app:comparison}

\subsection{Evaluation of Previous Methods}
In \autoref{sec:comparison}, we compare our work to previous NVS methods.
While we mostly follow the evaluation strategy outlined in GenWarp~\cite{seo2024genwarp}, their use of data filtering and the lack of explicit data sampling procedures prevent us from performing the exact same evaluation as in previously published results.
For this reason, we employ our own source and destination view procedures for evaluation (as detailed in \autoref{app:implementation}). We use the officially published code for evaluating GeoGPT~\cite{rombach2021geometry}, PhotoNVS~\cite{yu2023long}, and GenWarp~\cite{seo2024genwarp}.
All models were evaluated on the $256\times256$ resolution. 
Because GeoGPT and GenWarp were trained to operate in different resolutions ($208\times368$ and $512\times512$ respectively) we resize the source center-cropped $360\times360$ images from RealEstate10K~\cite{realestate10k} to the relevant resolution before the NVS operation, and then resize again to $256\times256$ to perform the evaluation.
Additionally, GeoGPT uses 3D points (based on COLMAP~\cite{schoenberger2016sfm, schoenberger2016mvs} from the entire scene) to scale its estimated depth input and solve the NVS task. This gives GeoGPT an unfair advantage other alternative methods for which the scale ambiguity problem remains.
For GenWarp, we used the publicly available checkpoints, which are not trained exclusively on RealEstate10K, as no RealEstate10K model is publicly available.

\subsection{Additional Metrics for Method Comparison}
We augment the comparison in \autoref{tab:comp} with 3 additional metrics:
(i) Joint FID~\cite{devries2019evaluation} (JFID), which is a Fréchet distance over concatenated features from both the source and destination (ground-truth or generated) images; 
(ii) Fréchet distance computed in the DINOv2~\cite{oquab2023dinov2} space following~\cite{karras2024analyzing, stein2023exposing}, named FDD (Fréchet DINOv2 Distance);
and (iii) JFDD (Joint Fréchet DINOv2 Distance), which is the same as JFID but using DINOv2 instead of Inception.
The full results are shown in \autoref{tab:apdx}.
The source image is included in the table to provide a frame of reference for the different metrics.
We expect the source image to achieve high perceptual quality in the metrics, because the source and destination images are both drawn from the same dataset, and thus in theory they should have the same probability distribution.\footnote{FID is limited here to 10K images, creating a discrepancy between source and destination views.}
At the same time, the PSNR of the source images is consistently subpar compared to NVS methods, as (obviously) the source views do not solve the NVS task.
To quantify the perceptual quality of a generated sample that does solve the NVS task we propose using JFID~\cite{devries2019evaluation}, a metric that attempts to measure similarity of conditional distributions.
The source image under-performs all NVS methods in JFID as expected.
Our method (named ``VIVID (RE10K)'' in the table) is superior to all the tested methods in JFID, in line with the improvement in both FID and PSNR, making it state-of-the-art across all metrics in both mid- and long- range.
The results shown in FDD and JFDD corroborate our conclusions, and provide more robust perceptual quality metrics~\cite{stein2023exposing}.
Finally, we also include the version of our model trained on both RealEstate10K and OpenImages as ``VIVID (RE10K + OI)'' in the table, showing that training on single-view data with our proposed $10\%$ augmentation does not degrade performance on RealEstate10K, and even offers a slight improvement.

\section{Additional Generalization Examples}
\begin{figure*}
    \centering
    \includegraphics[width=\linewidth]{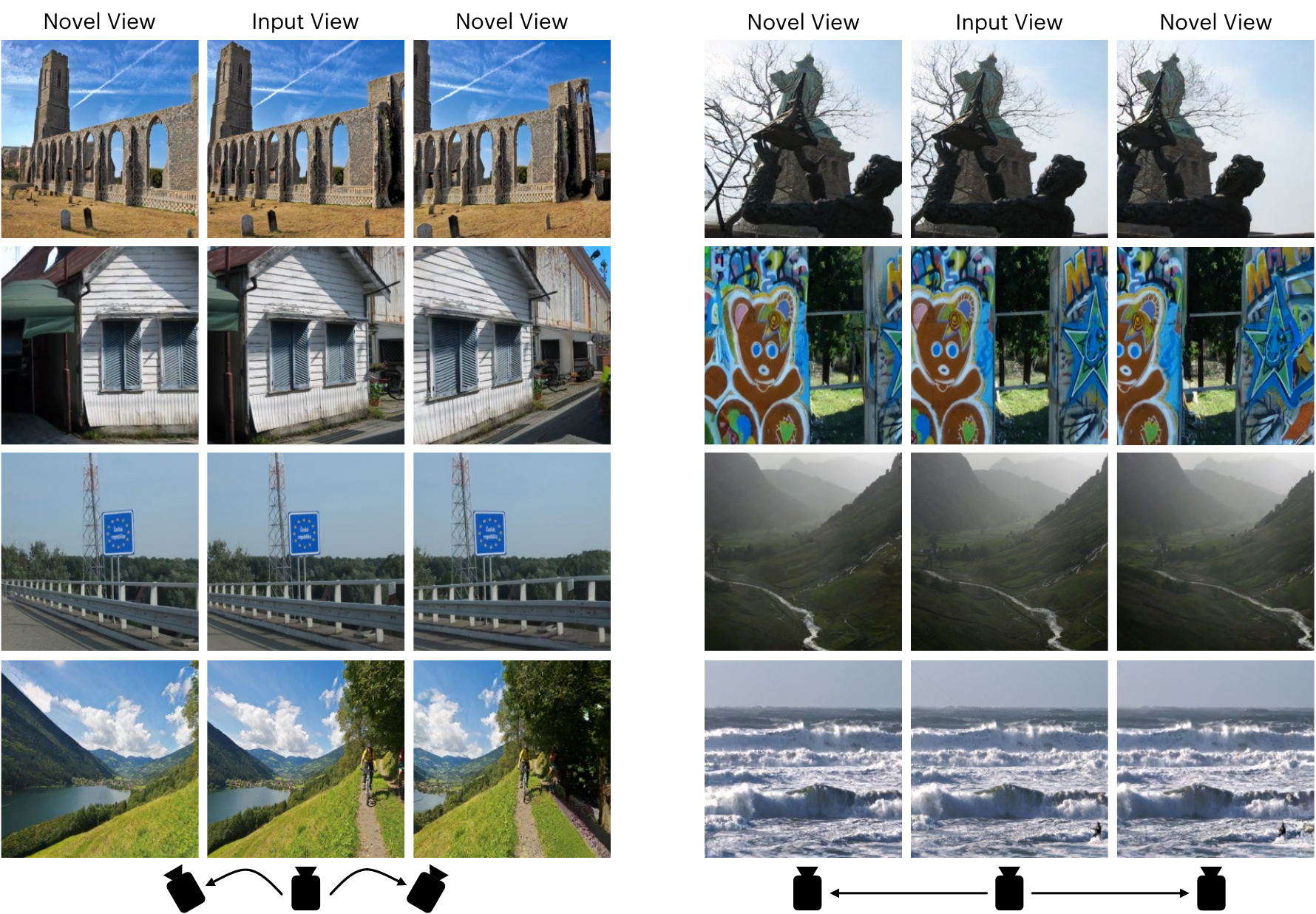}
    \caption{Additional examples for generalization from the OpenImages dataset, exhibiting rotation (left) and translation (right).}
    \label{fig:additional-generalization}
\end{figure*}

\autoref{fig:additional-generalization} shows additional qualitative examples for generalization with our model on images from OpenImages. 

\begin{table*}
  \centering
  \begin{tabular}{@{}l|ccccc|ccccc@{}}
    \toprule
    &  \multicolumn{5}{c}{Mid-range} & \multicolumn{5}{c}{Long-range} \\
    Method & FID $\downarrow$ & PSNR $\uparrow$ 
    & JFID $\downarrow$ & FDD $\downarrow$ & JFDD $\downarrow$ 
    & FID $\downarrow$ & PSNR $\uparrow$ & JFID $\downarrow$ & FDD $\downarrow$ & JFDD $\downarrow$ \\
    
    \midrule
    GeoGPT~\cite{rombach2021geometry} & $6.43$ & $14.06$ 
    &$13.19$ & $288.26$ & $444.29$
    & $7.22$ & $13.13$ & $13.46$ & $332.49$ & $455.78$

    \\
    PhotoNVS~\cite{yu2023long} & $7.12$ & $13.32$ 
    & $13.62$ & $433.47$& $559.42$
    & $9.22$ & $12.05$ & $15.76$ & $552.42$ & $668.88$ \\

    GenWarp~\cite{seo2024genwarp} & $5.91$ & $13.43$
     & $10.52$ & $68.70$ & $101.39$
    & $7.38$ & $12.10$ 
    & $13.62$ & $120.54$ & $171.27$
    \\
    VIVID (RE10K) & $\underline{2.89}$ & $\underline{17.36}$ 
    & $\underline{6.26}$ & $\underline{41.20}$ & $\underline{67.43}$
    & $\underline{3.89}$ & $\textbf{15.21}$ 
    & $\underline{8.18}$ & $\underline{80.44}$ & $\underline{118.75}$
    \\
    VIVID (RE10K + OI) & $\textbf{2.82}$ & $\textbf{17.38}$ 
    & $\textbf{6.14}$ & $\textbf{38.13}$ & $\textbf{63.37}$
    & $\textbf{3.77}$ & $\underline{15.19}$ 
    & $\textbf{8.02}$ & $\textbf{72.62}$ & $\textbf{109.53}$
    \\
    \midrule
    Source Image & $2.58$ & $13.12$ 
     & $73.76$ & $19.32$ & $528.71$
    & $3.00$ & $11.91$ 
     & $51.07$ & $14.09$ & $309.47$
    \\

    \bottomrule
  \end{tabular}
  \caption{Comparison to previous methods, including additional metrics. Evaluation is done on ${10}$K source-target pairs from RealEstate10K. Best results in each column are in bold, second best are underlined.}
  \label{tab:apdx}
\end{table*}

\end{document}